\documentclass[sigconf,nonacm]{acmart}

\usepackage{booktabs}
\usepackage{multirow}
\usepackage{graphicx}
\usepackage{subcaption}

\begin{document}

\title{Sparse Autoencoders Reveal Interpretable Egocentric State-Change Features in Frozen V-JEPA Representations}

\author{Eduardus Tjitrahardja}
\affiliation{%
  \institution{Tsinghua University}
  \city{Beijing}
  \country{China}
}
\email{xhp25@mails.tsinghua.edu.cn}

\begin{abstract}
Self-supervised video encoders produce transferable spatiotemporal
representations, but their internal features remain difficult to inspect.
We train a Top-\(K\) sparse autoencoder (SAE) on frozen V-JEPA 2 token
representations from 5,000 balanced clips sampled from Ego4D's Object
State Change Classification benchmark. We evaluate the resulting sparse
codes using linear probes and PNR-grounded qualitative analyses. 
SAE t32 achieves 0.751 AUROC and 0.682 balanced accuracy on one video-disjoint validation split, compared
with 0.702 and 0.654 for the corresponding dense representation. Each
token activates 32 of 16,384 latents, while 7.1\% of latents fire at
least once per clip on average. Among the ten highest positive
PNR-window-contrast latents, five receive tentative recurring visual
descriptions from five reviewed clips each, including hand-held object
repositioning and hand-applied force. Peak-activation timing is only
weakly associated with PNR location (\(\max |\rho|=0.15\)). These
results suggest that sparse decomposition can preserve
linearly decodable state-change information while surfacing candidate
hand--object features for further auditing.
\end{abstract}



\maketitle

\section{Introduction}

Self-supervised video representation models~\cite{bardes2024vjepa,feichtenhofer2022masked,tong2022videomae} learn transferable spatiotemporal features from unlabeled video. These developments build on broader advances in self-supervised representation learning~\cite{chen2020simple,he2020momentum} and transformer-based visual architectures~\cite{dosovitskiy2021image}, extending representation learning from static images to temporal visual sequences. V-JEPA (Video Joint-Embedding Predictive Architecture)~\cite{bardes2024vjepa} learns by predicting representations of masked spatiotemporal regions rather than reconstructing their pixels, and performs strongly on action-recognition benchmarks including Kinetics-400~\cite{carreira2017quo} and Something-Something-v2~\cite{goyal2017something}. V-JEPA 2~\cite{assran2025vjepa2} scales this latent-prediction approach to substantially larger models and pretraining data, producing strong representations for motion understanding and action anticipation. Nevertheless, it remains unclear how frozen V-JEPA 2 representations encode object state changes in egocentric video. Inspectable representations could help determine whether downstream predictions rely on hand--object interaction evidence, object appearance, scene context, or temporal shortcuts.

Mechanistic interpretability seeks to identify internal representations and computations that contribute to model behavior~\cite{olah2020zoom,elhage2022superposition}. Sparse autoencoders (SAEs) learn overcomplete dictionaries that reconstruct dense model activations using sparse latent codes~\cite{cunningham2023sparse,gao2024scaling}. Studies of language models have found that many SAE latents admit coherent human interpretations and can isolate features that are obscured within individual model dimensions~\cite{bricken2023monosemanticity,templeton2024scaling,gao2024scaling}. SAE-based analysis has also begun to extend to vision and video representations~\cite{stevens2025saevision,dokme2026spatiotemporal}. However, to our knowledge, prior work has not examined sparse features in frozen V-JEPA 2 representations for egocentric object state changes or related their temporal activations to annotated state-change onset. Video representations introduce additional interpretability challenges because individual features may depend jointly on spatial appearance, object motion, hand--object interactions, and temporal context.

We apply a Top-(K) SAE to frozen V-JEPA 2~\cite{assran2025vjepa2} features extracted from clips in Ego4D's Object State Change Classification (OSCC) benchmark~\cite{grauman2022ego4d}. Our contributions are:

\begin{itemize}
\item \textbf{Sparse feature decomposition:} We train a Top-(K) SAE on frozen V-JEPA 2 features, producing a sparse latent dictionary for inspecting evidence related to object state changes.

\item \textbf{Task-level evaluation:} We compare dense V-JEPA 2 representations and SAE-derived representations using linear probes for binary object state-change classification.

\item \textbf{Latent inspection:} We inspect selected SAE latents through PNR-grounded high-activation clips and retain manual descriptions only when an observable pattern recurs across reviewed examples.

\item \textbf{Temporal analysis:} For state-change-positive clips, we compare latent activation profiles with PNR annotations to characterize whether individual latents concentrate near the annotated onset of a state change or capture broader evidence distributed across the clip.
\end{itemize}

Together, these analyses provide a framework for auditing selected features within a self-supervised video representation and assessing their relationship to object state-change evidence.

\section{Related Work}

\subsection{Self-Supervised Video Learning}

Self-supervised video models learn transferable spatiotemporal representations from unlabeled video. Masked approaches such as VideoMAE~\cite{tong2022videomae} and MAE-ST~\cite{feichtenhofer2022masked} reconstruct heavily masked video inputs, whereas V-JEPA~\cite{bardes2024vjepa} predicts target-encoder representations of masked spatiotemporal regions in latent space rather than reconstructing pixels. V-JEPA 2~\cite{assran2025vjepa2} scales this objective to larger models and pretraining data, achieving strong transfer on motion understanding and action anticipation. Related foundation models such as InternVideo~\cite{wang2022internvideo} combine generative and discriminative objectives for general video representation learning. Despite their strong downstream performance, the semantic and temporal organization of their internal features remains poorly understood.

\subsection{Mechanistic Interpretability and Sparse Autoencoders}

Mechanistic interpretability studies the internal representations and computations underlying model behavior~\cite{olah2020zoom}. One proposed challenge is superposition, in which multiple features are represented along overlapping directions~\cite{elhage2022superposition}. Sparse autoencoders (SAEs) address this problem by learning overcomplete dictionaries that reconstruct dense activations using sparse latent codes~\cite{cunningham2023sparse}. Large-scale studies have found that many SAE latents admit coherent human interpretations~\cite{templeton2024scaling}, while Top-\(K\) SAEs directly retain the \(k\) largest activations and improve the reconstruction--sparsity trade-off~\cite{gao2024scaling}. Recent work has extended SAEs to video representations~\cite{dokme2026spatiotemporal}; however, sparse features in frozen V-JEPA 2 representations have not been studied for egocentric object state changes or compared temporally with PNR annotations.

\subsection{Egocentric Video and State-Change Understanding}

Egocentric video records visual experience from a camera wearer's first-person perspective and commonly captures hand--object interactions and procedural activities~\cite{damen2018scaling,grauman2022ego4d}. 
EPIC-KITCHENS~\cite{damen2018scaling} established a large-scale benchmark for egocentric action recognition and anticipation in kitchen environments. Ego4D~\cite{grauman2022ego4d} broadened this setting with 3,670 hours of video and five benchmark suites covering episodic memory, hands and objects, audio-visual diarization, social interaction, and forecasting.

Within Ego4D's Hands and Objects benchmark, an object state change denotes a physical change in properties such as size, shape, composition, or texture. Object State Change Classification (OSCC) predicts whether an object state change occurs within an short video clip, whereas Point-of-No-Return (PNR) temporal localization identifies the keyframe at which the state change begins~\cite{grauman2022ego4d}. 
Existing methods trained on these benchmarks have primarily targeted predictive performance, including divided space-time attention for OSCC~\cite{islam2022oscc}, joint VideoMAE fine-tuning for OSCC and PNR~\cite{lei2022masked}, and the transfer of pretrained egocentric video-language representations to OSCC~\cite{lin2022egovlp}. 
More recent cross-task approaches train or adapt models for OSCC alongside PNR and other Ego4D tasks~\cite{xue2023egot2,peirone2024egopack}. 
Although some of these approaches analyze cross-task attention or task prototypes, they do not investigate sparse, individually interpretable features within a frozen video representation. In contrast, we train sparse autoencoders on frozen V-JEPA representations and use PNR annotations to examine whether learned feature activations align temporally with the annotated onset of object state changes.

\subsection{Linear Probing and Representation Analysis}

Linear probes evaluate frozen representations by measuring how well task-relevant information can be decoded with a simple linear classifier~\cite{alain2017understanding}. We use linear probes to compare dense V-JEPA 2 representations with SAE-derived sparse representations on OSCC, testing how much linearly decodable state-change information is preserved after sparse decomposition.

\section{Method}

\begin{figure*}[t]
\centering
\includegraphics[width=\textwidth]{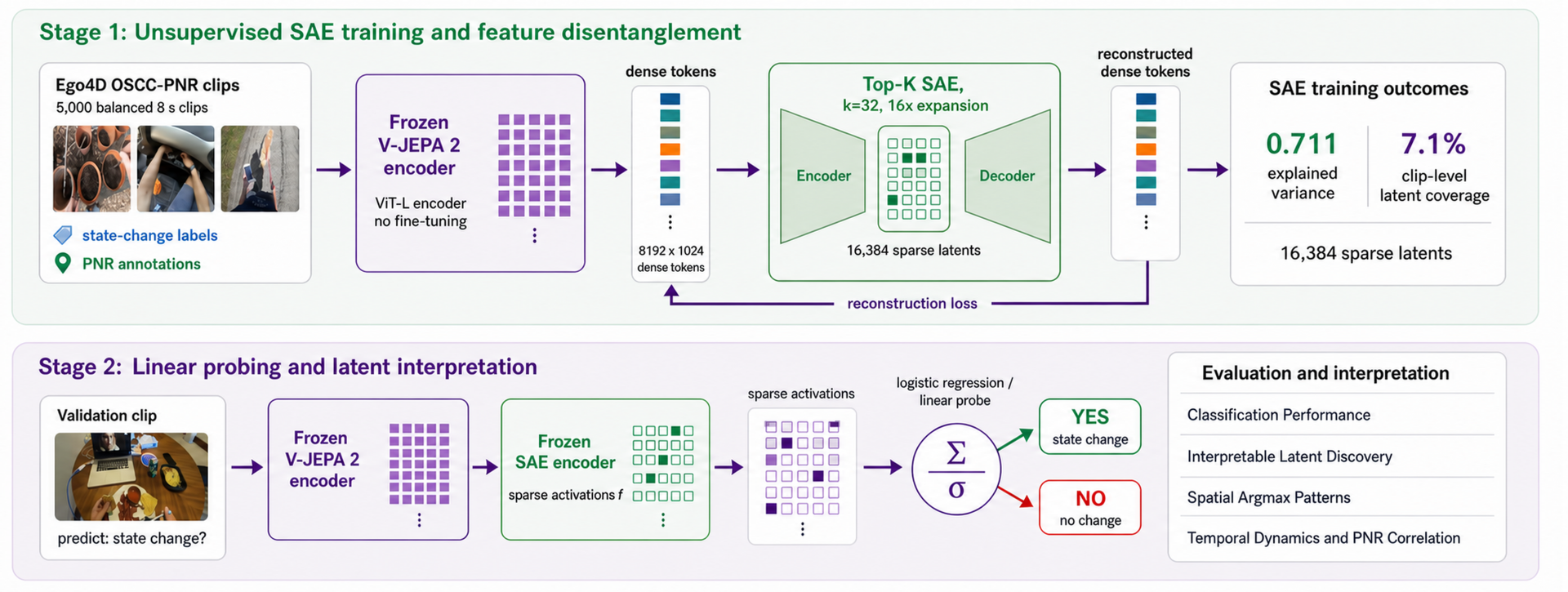}
\caption{Overview of SAE extraction, evaluation, and interpretability analysis.}
\Description{A wide diagram from 5,000 Ego4D OSCC-PNR clips through frozen V-JEPA 2 and a Top-K SAE to training metrics. Four lower panels show t32 linear-probe results, PNR-window contrast, a 16 by 16 spatial coordinate-frequency grid with a hand clip, and 32-bin temporal diagnostics.}
\label{fig:pipeline}
\end{figure*}

Figure~\ref{fig:pipeline} provides an overview of the pipeline, from balanced OSCC-PNR clip sampling and frozen V-JEPA 2 feature extraction to sparse decomposition, temporal pooling, linear-probe evaluation, and interpretability analysis.

\subsection{Dataset and Task}

We use the Ego4D FHO OSCC-PNR benchmark~\cite{grauman2022ego4d}. We filter the benchmark to instances with exactly 8\,s clip duration to ensure consistent temporal coverage, then sample 5,000 clips: 2,500 positive (state change occurs) and 2,500 negative (no state change), split by video ID to prevent data leakage (3,000 train, 2,000 val; 90 train videos, 60 val videos). Each retained clip is 8 seconds at 30 FPS (240 frames). Positive clips include PNR frame annotations. The task is binary classification: given a clip, predict whether a state change occurs.

\subsection{V-JEPA Feature Extraction}

We use the frozen V-JEPA 2 ViT-L checkpoint \path{facebook/vjepa2-vitl-fpc64-256}~\cite{assran2025vjepa2} (FPC64 at 256 pixels) with no fine-tuning. We sample 64 frames uniformly from each 240-frame clip and resize frames to $256 \times 256$. We then pass these frames to the V-JEPA 2 video processor for model-specific preprocessing before feature extraction. V-JEPA tokenizes video into 2-frame tubelets with $16 \times 16$ spatial patches, yielding a feature tensor of shape $[32 \text{ temporal} \times 256 \text{ spatial} \times 1024 \text{ hidden}] = [8192, 1024]$.

\subsection{Sparse Autoencoder Training}

We train a Top-K SAE~\cite{gao2024scaling} to decompose V-JEPA features into sparse latent activations. The encoder applies per-dimension standardization, a linear map, ReLU, and Top-$K$ gating:

\begin{equation}
\tilde{\mathbf{x}} = \frac{\mathbf{x}-\boldsymbol{\mu}}{\boldsymbol{\sigma}}, \qquad
\mathbf{h} = \operatorname{ReLU}(W_e\tilde{\mathbf{x}}+\mathbf{b}_e), \qquad
\mathbf{z} = \operatorname{TopK}_k(\mathbf{h})
\end{equation}
\begin{equation}
\hat{\mathbf{x}} = \boldsymbol{\sigma} \odot (W_d\mathbf{z}+\mathbf{b}_d)+\boldsymbol{\mu}.
\end{equation}

Here $\mathbf{x} \in \mathbb{R}^{1024}$ is a V-JEPA token, $\boldsymbol{\mu}$ and $\boldsymbol{\sigma}$ are per-dimension normalization statistics, $W_e \in \mathbb{R}^{16384 \times 1024}$ is the encoder weight matrix, and $W_d \in \mathbb{R}^{1024 \times 16384}$ is the decoder weight matrix. $\operatorname{TopK}_k$ retains the $k=32$ largest nonnegative activations and sets the remaining entries to zero. This yields a 16$\times$ expansion (1,024 $\to$ 16,384 latents) with at most 32 nonzero latents per token.

We train on all training-set V-JEPA tokens (3,000 clips $\times$ 8,192 tokens/clip $= 24,576,000$ tokens per epoch) for 10 epochs using AdamW with learning rate $10^{-4}$. The normalization statistics are estimated from a random 200K-token subset; the estimated variance is clamped to a minimum of $10^{-6}$ before square-rooting, so the resulting standard deviations are at least $10^{-3}$. After each optimizer update, we normalize the decoder columns to unit $\ell_2$ norm. The reconstruction loss is the mean squared error:

\begin{equation}
\mathcal{L} = \frac{1}{BD}\sum_{i=1}^{B}\sum_{j=1}^{D}(x_{ij}-\hat{x}_{ij})^2,
\end{equation}
where $B$ is the token-batch size and $D=1024$ is the input dimension.

Final metrics: MSE = 3.095, explained variance = 0.711, and a mean clip-level active fraction of 7.09\% (1,162 of 16,384 distinct latents active per clip). No latent had zero firing rate in the final training epoch. Appendix Table~\ref{tab:sae-training-epochs} reports the epoch-wise training trajectory, and Appendix~\ref{apdx:hyperparam_training_details} gives the full SAE and linear-probe configuration. During pooled-feature extraction, the encoder path re-applies normalization, ReLU, and Top-$K$ gating to each token; the decoder reconstruction is not used.

\subsection{Temporal Pooling}

V-JEPA features have 32 temporal bins (2 frames per bin). We extract SAE activations for each token, then pool across spatial and temporal dimensions. We denote pooling strategies as t$N$, where $N$ is the number of temporal bins retained after pooling. We compare two strategies:

\begin{itemize}
\item \textbf{t8 pooling:} Group 32 bins into 8 coarse bins (8 frames each), max-pool spatially within each bin. Output: $[8 \times 16384] = 131,072$ features per clip.
\item \textbf{t32 pooling:} Keep all 32 bins (2 frames each), max-pool spatially within each bin. Output: $[32 \times 16384] = 524,288$ features per clip. This preserves tubelet-level resolution and aligns with V-JEPA's native temporal granularity.
\end{itemize}

The t32 representation preserves V-JEPA's native two-frame tubelet resolution, whereas t8 is a coarser temporal ablation. The t8/t32 comparison tests whether this additional temporal resolution benefits state-change classification.

\subsection{Linear Probes and Baselines}

We train linear classifiers on pooled features to evaluate task performance:

\begin{itemize}
\item \textbf{Dense V-JEPA:} Pool raw V-JEPA features (no SAE). We test global mean, global max, t8, and t32 pooling. Classifier: sklearn logistic regression with balanced class weights.
\item \textbf{SAE features:} Pool SAE activations with t8 or t32. Classifier: single-layer PyTorch linear classifier trained for 50 epochs with learning rate $10^{-3}$.
\end{itemize}

Metrics: balanced accuracy (mean of per-class recall) and AUROC.

\subsection{Interpretability Analysis}

We analyze SAE latents through:

\begin{enumerate}
\item \textbf{Activation contrast:} For each latent, compute mean activation inside a $\pm$2-bin window around the PNR frame versus mean activation elsewhere in positive clips. Latents with high positive contrast are candidate state-change activators; latents with high negative contrast are suppressors near PNR.

\item \textbf{PNR-grounded manual review:} For each candidate latent, select five positive validation clips. Activator examples peak inside the $\pm$2-bin PNR window; suppressor examples peak outside it. A qualitative observation is retained only when a visible pattern recurs in five positive sampled clips.

\item \textbf{Spatial argmax patterns:} For selected retained activators, we use both clip-level and aggregate diagnostics. For a selected PNR-grounded review clip, we show the peak-bin spatial argmax and an activation-weighted argmax map over all 32 bins. For the aggregate diagnostic, we collect the spatial max-pooling argmax at each clip's highest t32 activation and aggregate activation-weighted patch frequencies over the 64 highest-activation validation clips.

\item \textbf{Temporal profiles:} Average latent activation across 32 bins for positive clips and compare it descriptively with PNR timing.

\item \textbf{PNR correlation:} For positive clips with valid PNR annotations, compute Spearman correlation between each latent's peak-activation bin and the PNR bin.
\end{enumerate}

\section{Results}

\subsection{Classification Performance}

Table~\ref{tab:probes} reports linear-probe performance on the Ego4D OSCC validation split. Temporal pooling consistently outperforms global pooling for the dense V-JEPA 2 representation, with t32 achieving 0.654 balanced accuracy and 0.702 AUROC. Applying the SAE further improves the observed performance: SAE t32 obtains the best results on this split, reaching 0.682 balanced accuracy and 0.751 AUROC. For both dense and SAE representations, t32 outperforms t8, suggesting that preserving finer temporal resolution is useful for state-change classification. We therefore use t32 features in the subsequent temporal and interpretability analyses, as they provide the strongest validation performance while retaining all 32 native temporal bins.

\begin{table}[h]
\centering
\caption{Linear-probe performance for binary state-change classification on the Ego4D OSCC validation split.}
\label{tab:probes}
\begin{tabular}{lccc}
\toprule
\textbf{Method} & \textbf{Pooling} & \textbf{Bal. Acc.} & \textbf{AUROC} \\
\midrule
V-JEPA Dense & global mean & 0.575 & 0.623 \\
V-JEPA Dense & global max & 0.575 & 0.600 \\
V-JEPA Dense & t8 & 0.613 & 0.665 \\
V-JEPA Dense & t32 & 0.654 & 0.702 \\
\midrule
SAE (ours) & t8 & 0.650 & 0.709 \\
SAE (ours) & t32 & \textbf{0.682} & \textbf{0.751} \\
\bottomrule
\end{tabular}
\begin{minipage}{\columnwidth}
\footnotesize\textit{Note.} Results are reported for a single split and are not averaged across random seeds or cross-validation folds.
\end{minipage}
\end{table}

The SAE representation combines competitive task performance with a sparse feature basis suitable for latent-level inspection. Each clip activates an average of 1,162 out of 16,384 SAE latents, corresponding to 7.1\% clip-level sparsity (Appendix Table~\ref{tab:sae-training-epochs}). In comparison, the original V-JEPA 2 representation contains 1,024 dense feature dimensions. These results indicate that the sparse decomposition preserves substantial linearly decodable state-change information while providing a higher-dimensional set of features for subsequent interpretation.

\subsection{Interpretable Latent Discovery}

We identify latents with high activation contrast around the annotated PNR window. For each latent, contrast is the mean activation in the $\pm$2-bin PNR window minus the mean activation elsewhere, computed over 1,000 positive validation clips. Table~\ref{tab:latents} shows the five strongest activators (higher activation near PNR) and five strongest suppressors (lower activation near PNR); the complete 20-latent (10 each) table and additional review examples appear in Appendix~\ref{apdx:complete_latent_material}.

\begin{table}[h]
\centering
\caption{Representative SAE latents by PNR-window activation contrast.}
\label{tab:latents}
\begin{tabular}{lcc}
\toprule
\textbf{Latent ID} & \textbf{Type} & \textbf{PNR Contrast} \\
\midrule
4152 & Activator & 1.37 \\
14205 & Activator & 1.06 \\
3610 & Activator & 0.99 \\
12848 & Activator & 0.94 \\
12826 & Activator & 0.90 \\
\midrule
5330 & Suppressor & $-1.76$ \\
2971 & Suppressor & $-1.42$ \\
9921 & Suppressor & $-1.30$ \\
3546 & Suppressor & $-0.72$ \\
6178 & Suppressor & $-0.63$ \\
\bottomrule
\end{tabular}
\end{table}

\begin{figure}[h]
\centering
\includegraphics[width=0.48\textwidth]{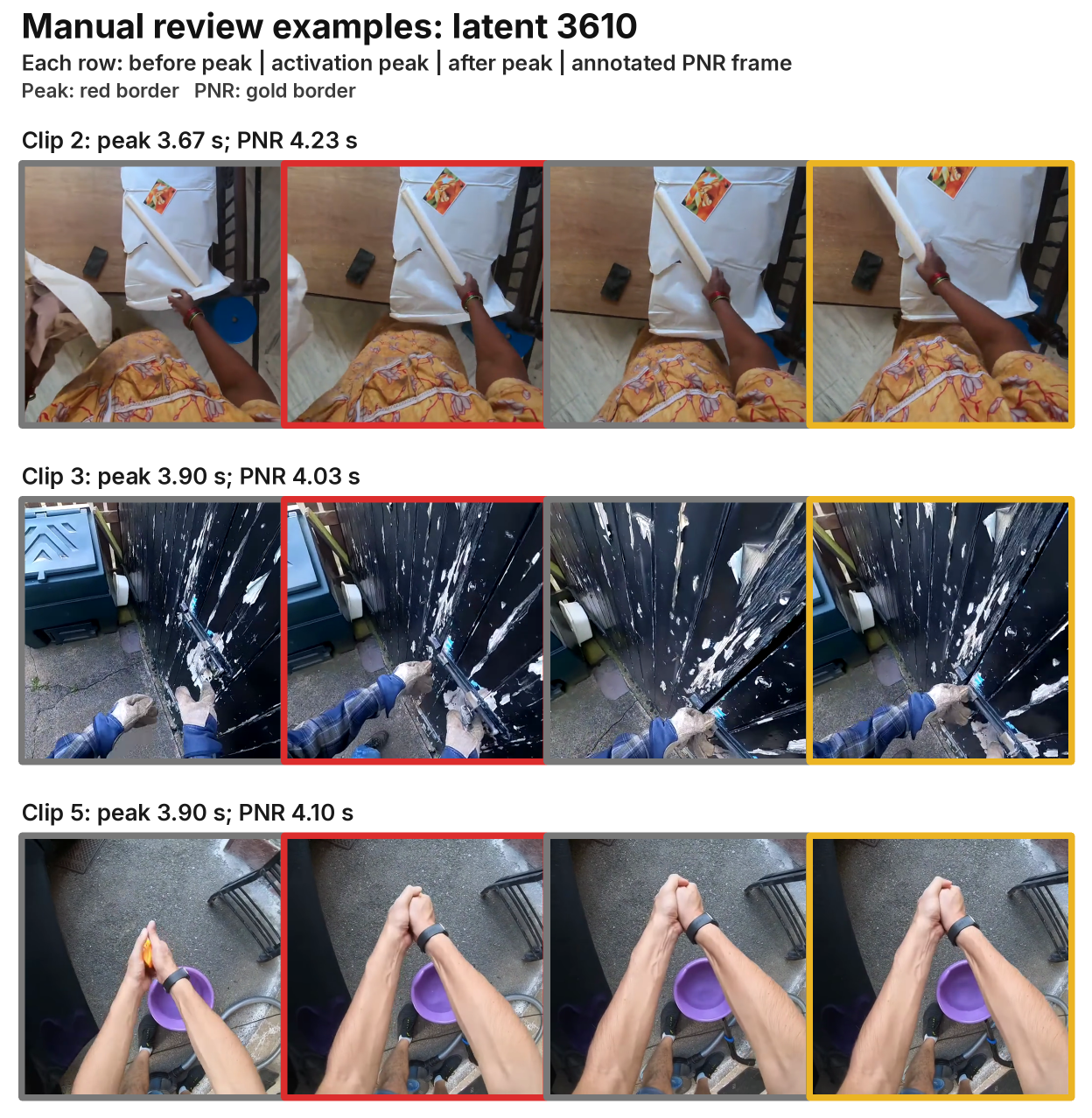}
\caption{PNR-grounded manual-review examples for activator latent 3610.}
\Description{Three egocentric video examples for SAE latent 3610. Each row shows frames before, at, and after the selected activation peak, followed by the annotated PNR frame.}
\label{fig:manual-review}
\end{figure}

For qualitative inspection, we assessed five sampled positive clips for each activator and suppressor. Five activators showed recurring visual patterns, while the other five were marked ambiguous. The clearest activator pattern was latent 3610, whose high activations co-occurred with a hand actively changing its hold on or repositioning a nearby object. Across suppressors, frames around the selected outside-PNR peaks typically show a locally sustained activity or viewpoint: the user continues an ongoing task without an abrupt local visual change. We report this as a shared temporal-context observation, rather than an action or object label. Figure~\ref{fig:manual-review} shows three clips example of activator latent 3610, and Appendix Figure~\ref{fig:suppressor-review} gives analogous examples for suppressor latent 5330.

\subsection{Spatial Argmax Patterns}

\begin{figure}[h]
\centering
\includegraphics[width=0.486\textwidth]{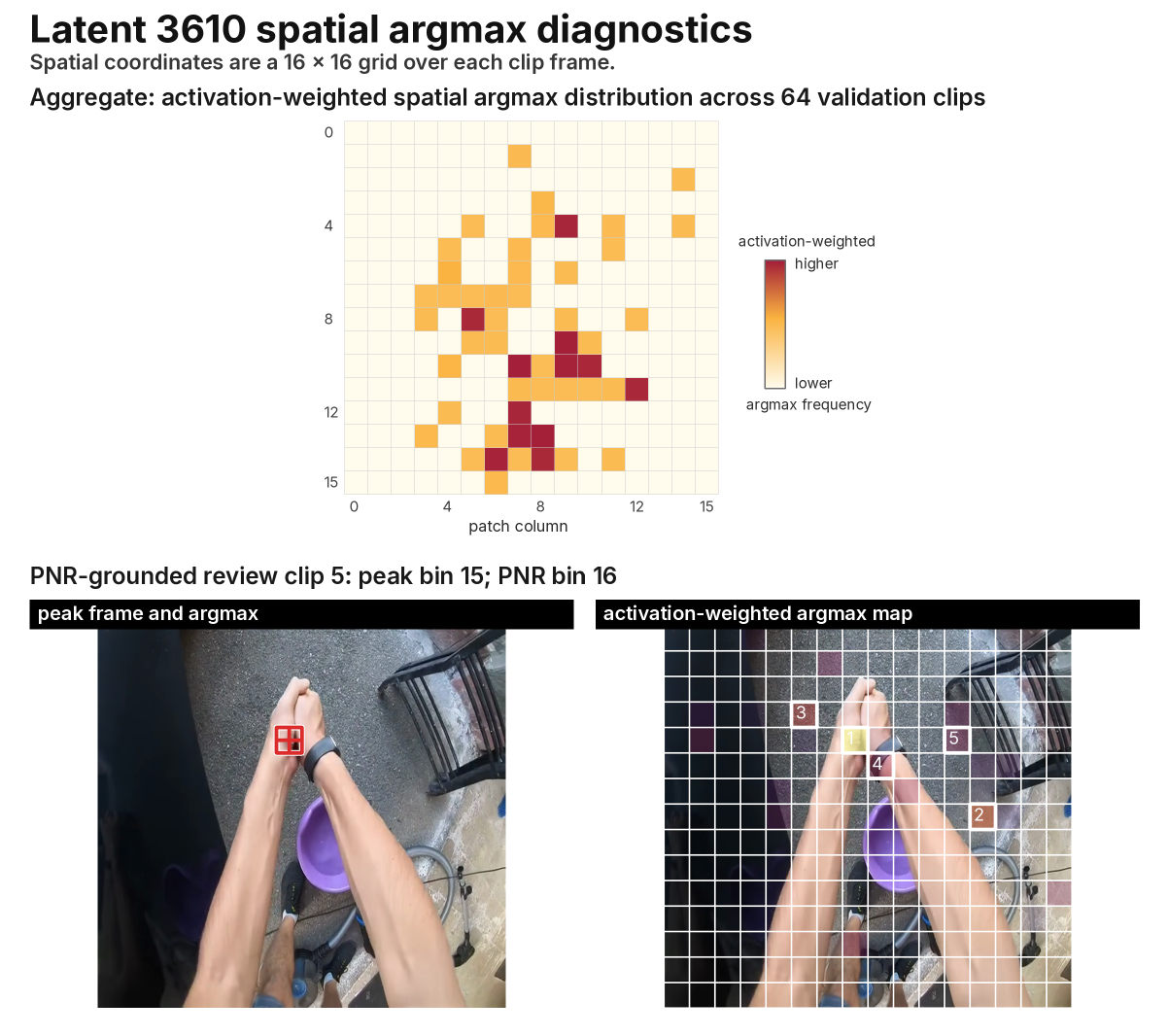}
\caption{Aggregate and clip-level spatial-argmax diagnostics for activator latent 3610.}
\Description{A figure for latent 3610. The top row is a standalone activation-weighted 16 by 16 spatial-argmax frequency grid over 64 validation clips. The bottom row shows the fifth manual-review clip, with the peak-frame argmax on a visible hand and an activation-weighted argmax map across its 32 temporal bins.}
\label{fig:spatial}
\end{figure}

Figure~\ref{fig:spatial} combines aggregate and clip-level spatial-argmax diagnostics for latent 3610. The top row is a standalone 16 $\times$ 16 coordinate-frequency grid: for each of the 64 highest-activation validation clips, we select the highest activation over all 32 bins and add its activation weight at the corresponding spatial argmax position. The resulting central concentration is consistent with the image region in which hand manipulation is commonly visible in an egocentric frame; it is an image-coordinate trend, not hand or object localization. The bottom row shows the fifth PNR-grounded review clip: its peak is in bin 15, immediately before the PNR bin 16, and its spatial argmax patch lies on the visible hand. Its activation-weighted map across all 32 bins also places high weight in the hand--object work area. This clip-level diagnostic is consistent with the repeated hand-held object repositioning observation.

\subsection{Temporal Dynamics and PNR Correlation}

\begin{figure*}[t]
\centering
\includegraphics[width=\textwidth]{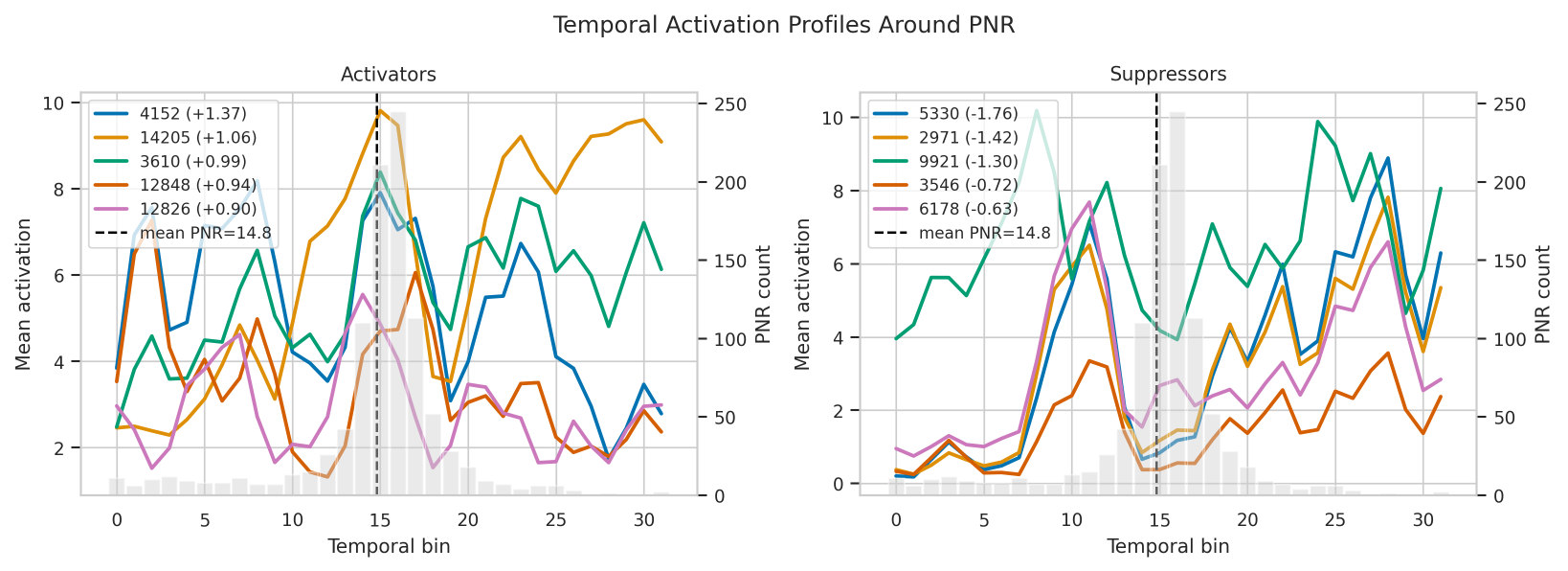}
\caption{Temporal activation profiles for top SAE activators and suppressors relative to PNR timing.}
\Description{Two temporal activation line charts comparing activator and suppressor SAE latent responses across 32 temporal bins with PNR distribution overlays.}
\label{fig:temporal}
\end{figure*}

Figure~\ref{fig:temporal} shows mean activation across 32 temporal bins for the five highest-contrast activators and suppressors, averaged over positive validation clips with valid PNR frames. Gray bars give the PNR count in every bin, and the dashed line marks the mean PNR bin (14.8). The profiles situate each selected latent relative to annotated timing and show whether its response accumulates before, around, or after PNR.

Direct PNR position correlation is weak. Table~\ref{tab:pnr} shows the top latents by absolute Spearman correlation between peak-activation bin and the clip-relative PNR bin, computed over 1,000 positive validation clips with valid PNR annotations. The largest absolute effect is $|r|=0.15$. Thus, the bin containing a latent's highest activation does not systematically coincide with the annotated PNR bin; peak timing alone does not explain the selected latents' responses.

\begin{table}[h]
\centering
\caption{Top SAE latents by absolute peak-bin PNR correlation on the Ego4D OSCC validation split.}
\label{tab:pnr}
\begin{tabular}{lcc}
\toprule
\textbf{Latent ID} & \textbf{Spearman $r$} & \textbf{$p$-value} \\
\midrule
14276 & -0.150 & $1.8\times10^{-6}$ \\
15295 & 0.148 & $2.7\times10^{-6}$ \\
14923 & -0.134 & $2.3\times10^{-5}$ \\
4140 & 0.133 & $2.3\times10^{-5}$ \\
6110 & 0.130 & $3.5\times10^{-5}$ \\
\bottomrule
\end{tabular}
\end{table}

\section{Discussion}

\subsection{Interpretability vs. Performance}

SAE features achieve the highest observed performance in Table~\ref{tab:probes} while providing inspectable latent activations. The 0.751 AUROC of SAE t32 exceeds the dense t32 baseline on this validation split, showing that sparse features retain competitive state-change information in this setting. The 7.1\% clip-level sparsity also provides a compact interface for downstream analysis.

\subsection{Semantic Features vs. Temporal Biases}

Weak PNR peak-bin correlation (Table~\ref{tab:pnr}) indicates that the reviewed activator latents are not primarily explained by a fixed clip-relative time position. Their high PNR-window contrast (Table~\ref{tab:latents}), the recurring hand--object patterns observed in the review clips (Figure~\ref{fig:manual-review}), and the central spatial-argmax tendency for latent 3610 (Figure~\ref{fig:spatial}) instead point to local visual configurations around state change. Thus, in this setting, the SAE analysis favors a spatial-feature account of state-change cues over a simple temporal-position account. The temporal profiles in Figure~\ref{fig:temporal} describe how these responses are distributed relative to PNR timing.

\subsection{Hand-Object Interactions and Visual Grounding}

The PNR-grounded review produced five activator descriptions. The most coherent example, latent 3610, recurred across five clips in which a hand changed its hold on or repositioned a nearby object (Figure~\ref{fig:manual-review}). The other retained descriptions are deliberately broad: nearby object/material manipulation, pre-contact hand approach, hand-applied object/material force, and downward hand-held object/tool manipulation. Together, these visually recurring patterns provide manually grounded evidence that a subset of SAE latents responds to hand--object configurations surrounding state change. This connects the identified sparse features to prior egocentric-vision work showing the importance of hand--object cues for activity understanding~\cite{nagarajan2020grounded,shan2020understanding}.

\subsection{Limitations}

\textbf{Manual qualitative labeling:} We manually inspected five PNR-grounded clips for each of ten activators and ten suppressors. Five activators retained descriptions, five activators were ambiguous, and suppressors shared only a non-semantic sustained-context observation. This small manual audit may miss feature diversity. More review clips per latent and detailed human-authored descriptions ground truth of the visual event around PNR would provide a stronger target for evaluating these interpretations against structured event annotations.

\textbf{Single dataset:} We validate on Ego4D OSCC. Cross-dataset validation on other egocentric or state-change benchmarks (e.g., EPIC-KITCHENS, Ego4D LTA, activity anticipation) remains to be tested.

\textbf{Frozen V-JEPA:} We analyze pretrained V-JEPA without fine-tuning. Fine-tuned models may learn different features; SAE applied to fine-tuned representations could reveal task-specific adaptations.

\textbf{SAE architecture:} We use a per-token Top-K SAE with fixed $k=32$. A spatiotemporal SAE that learns features jointly across spatial locations and time was not tested; it may better capture motion and temporal transitions relevant to state change. Adaptive sparsity or hierarchical SAE may also improve feature granularity.

\section{Conclusion}

We demonstrate that Sparse Autoencoders expose inspectable state-change features in frozen V-JEPA representations. SAE-derived features achieve 0.751 AUROC and 0.682 balanced accuracy on Ego4D state-change classification and 7.1\% clip-level sparsity on our validation split. We rank 20 latents by PNR-window contrast, retaining five activator descriptions and a shared sustained-context observation for suppressors. Five activator latents received recurring pattern descriptions centered on hand--object interaction, whereas peak timing was only weakly associated with PNR location.

These findings motivate further work that evaluates the visual interpretations against detailed human-authored PNR event descriptions, tests spatiotemporal SAEs for jointly modeling movement and appearance, performs causal latent-ablation tests, and extends the analysis across egocentric state-change datasets.

We presented an exploratory analysis of frozen V-JEPA 2 representations
using a Top-\(K\) sparse autoencoder trained on Ego4D object
state-change clips. The sparse representation retained substantial
linearly decodable OSCC information and provided individual latents that
could be inspected through high-activation clips, spatial locations, and
PNR-relative activation patterns. Five positive-contrast latents received
tentative recurring descriptions centered on hand--object interaction,
whereas peak timing was only weakly associated with PNR location.

These results support SAEs as a useful interface for auditing video
representations, but they do not yet establish monosemantic
state-change detectors. Controlled probe comparisons, PNR-shuffled
temporal baselines, larger blinded human evaluations, causal latent
interventions, and spatiotemporal SAE architectures are needed to test
whether the identified features reliably and causally represent object
state changes.

\bibliographystyle{ACM-Reference-Format}
\bibliography{references}

\clearpage

\appendix

\section{Supplementary Material}

\subsection{SAE Training Trajectory}

Table~\ref{tab:sae-training-epochs} reports the complete training log for the Top-$K$ SAE.

\begin{table}[h]
\centering
\caption{Epoch-wise Top-$K$ SAE training metrics. Each epoch processes 24,576,000 training tokens. MSE and explained variance are averaged over token minibatches; clip $L_0$ counts distinct active latents across all tokens in a clip.}
\label{tab:sae-training-epochs}
\small
\begin{tabular}{rrrrrrr}
\toprule
\textbf{Epoch} & \textbf{MSE} & \textbf{Expl. Var.} & \textbf{$L_0$/token} & \textbf{Clip $L_0$} & \textbf{Clip \%} & \textbf{Dead} \\
\midrule
1  & 3.384 & 0.6846 & 32.0 & 1591 & 9.71 & 0 \\
2  & 3.152 & 0.7061 & 32.0 & 1394 & 8.51 & 0 \\
3  & 3.128 & 0.7084 & 32.0 & 1312 & 8.01 & 0 \\
4  & 3.119 & 0.7092 & 32.0 & 1263 & 7.71 & 0 \\
5  & 3.113 & 0.7098 & 32.0 & 1234 & 7.53 & 0 \\
6  & 3.109 & 0.7102 & 32.0 & 1209 & 7.38 & 0 \\
7  & 3.105 & 0.7106 & 32.0 & 1188 & 7.25 & 0 \\
8  & 3.101 & 0.7109 & 32.0 & 1180 & 7.20 & 0 \\
9  & 3.097 & 0.7113 & 32.0 & 1153 & 7.04 & 0 \\
10 & 3.095 & 0.7115 & 32.0 & 1162 & 7.09 & 0 \\
\bottomrule
\end{tabular}
\end{table}

\subsection{Complete Latent Tables and Additional Review Examples}
\label{apdx:complete_latent_material}

\begin{figure}[h]
\centering
\includegraphics[width=0.48\textwidth]{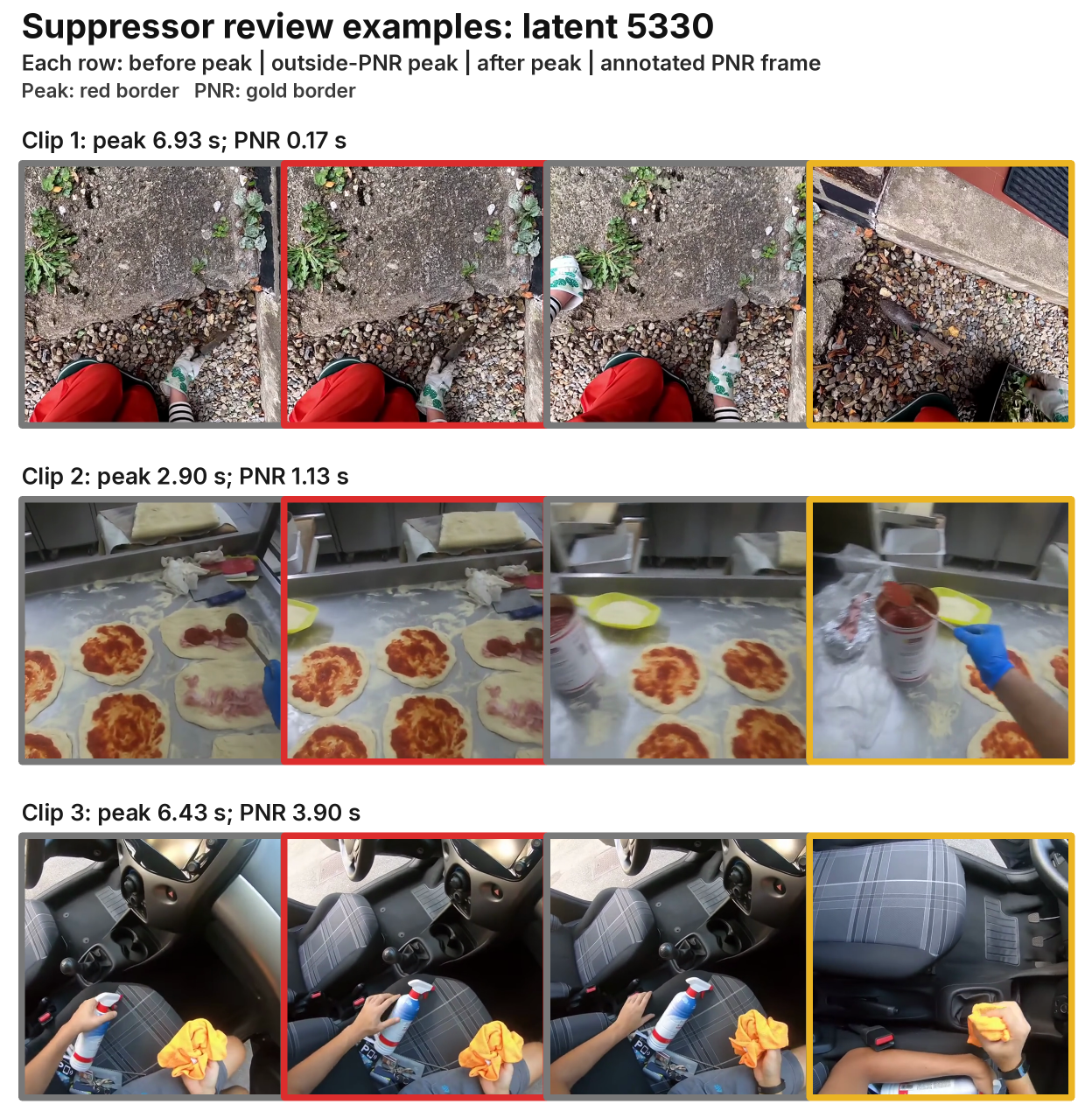}
\caption{PNR-grounded review examples for suppressor latent 5330.}
\Description{Three egocentric video sequences for suppressor latent 5330. Each row shows frames before, at, and after the selected outside-PNR activation peak, followed by the annotated PNR frame.}
\label{fig:suppressor-review}
\end{figure}

\begin{figure*}[t]
\centering
\includegraphics[width=0.96\textwidth]{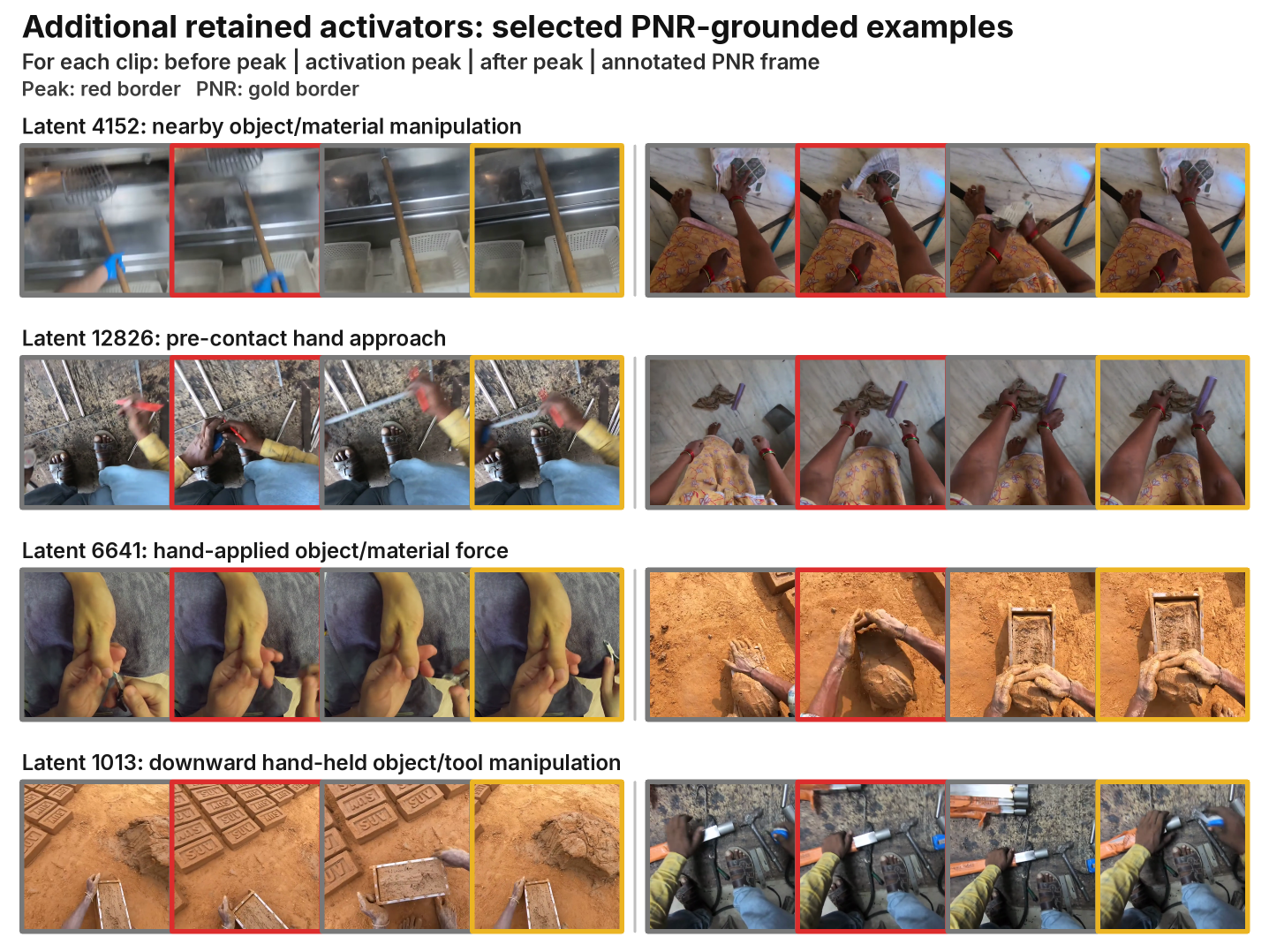}
\caption{Selected PNR-grounded examples for additional retained activators.}
\Description{Four rows of egocentric video frames for latents 4152, 12826, 6641, and 1013. Each row has two clips, with frames before, at, and after the activation peak followed by the annotated PNR frame.}
\label{fig:additional-activators}
\end{figure*}

Table~\ref{tab:all_latents} shows activation statistics for all 20 identified latents (10 activators, 10 suppressors). Figure~\ref{fig:additional-activators} shows PNR-grounded review examples for four additional retained activators. Figure~\ref{fig:suppressor-review} shows PNR-grounded review examples for suppressor latent 5330.

\begin{table}[h]
\centering
\caption{Complete contrast-ranked latent list and current manual-review outcome.}
\label{tab:all_latents}
\footnotesize
\begin{tabular}{lccp{3.2cm}}
\toprule
\textbf{Latent} & \textbf{Type} & \textbf{PNR Contrast} & \textbf{Review outcome} \\
\midrule
4152 & Activator & 1.37 & Nearby object/material manipulation \\
14205 & Activator & 1.06 & Ambiguous \\
3610 & Activator & 0.99 & Hand-held object repositioning \\
12848 & Activator & 0.94 & Ambiguous \\
12826 & Activator & 0.90 & Pre-contact hand approach \\
8617 & Activator & 0.90 & Ambiguous \\
6641 & Activator & 0.86 & Hand-applied object/material force \\
13741 & Activator & 0.80 & Ambiguous \\
7348 & Activator & 0.77 & Ambiguous \\
1013 & Activator & 0.69 & Downward hand-held object/tool manipulation \\
\midrule
5330 & Suppressor & -1.76 & Sustained local context \\
2971 & Suppressor & -1.42 & Sustained local context \\
9921 & Suppressor & -1.30 & Sustained local context \\
3546 & Suppressor & -0.72 & Sustained local context \\
6178 & Suppressor & -0.63 & Sustained local context \\
13976 & Suppressor & -0.62 & Sustained local context \\
9394 & Suppressor & -0.54 & Sustained local context \\
8883 & Suppressor & -0.52 & Sustained local context \\
3499 & Suppressor & -0.51 & Sustained local context \\
9975 & Suppressor & -0.50 & Sustained local context \\
\bottomrule
\end{tabular}
\end{table}

\subsection{Hyperparameters and Training Details}
\label{apdx:hyperparam_training_details}

\textbf{SAE Architecture:}
\begin{itemize}
\item Input dimension: 1,024 (V-JEPA hidden size)
\item Latent dimension: 16,384 (16$\times$ expansion)
\item Sparsity: Top-$k$ with $k=32$ (fixed per token)
\item Initialization: Decoder columns normalized to unit norm
\item Pre-decoder bias: Learned from 200K-token sample mean
\end{itemize}

\textbf{Training:}
\begin{itemize}
\item Optimizer: AdamW
\item Learning rate: $10^{-4}$ (no schedule)
\item Batch size: 256 tokens
\item Epochs: 10
\item Total tokens: 24,576,000 (3,000 clips $\times$ 8,192 tokens/clip)
\item Normalization: Z-score using 200K-token statistics
\end{itemize}

\textbf{Linear Probes:}
\begin{itemize}
\item Dense V-JEPA: sklearn LogisticRegression, balanced class weights, max iterations 1000
\item SAE features: PyTorch linear layer, AdamW optimizer, learning rate $10^{-3}$, 50 epochs, balanced BCEWithLogitsLoss
\end{itemize}

\end{document}